%% file: main.tex
\pdfoutput=1

\documentclass[11pt]{article}

\usepackage[preprint]{acl}

\usepackage{times}
\usepackage{latexsym}

\usepackage[T1]{fontenc}

\usepackage[utf8]{inputenc}

\usepackage{microtype}

\usepackage{inconsolata}

\usepackage{subcaption}
\usepackage{graphicx}
\usepackage{listings}
\usepackage[frozencache,cachedir=minted-cache]{minted}

\usepackage[acronym]{glossaries}
\newacronym{cot}{CoT}{chain-of-thought}
\newacronym{graddotinput}{$Grad \cdot Input$}{directional gradients}
\newacronym{gradnorm}{$GradNorm_{L2}$}{gradient norm}
\newacronym{ig}{IG}{integrated gradients}
\newacronym{is}{IS}{input salience}
\newacronym{it}{IT}{instruction-tuned}
\newacronym{lime}{LIME}{local interpretable model-agnostic explanations}
\newacronym{lit}{LIT}{Learning Interpretability Tool}
\newacronym{llm}{LLM}{large language model}
\newacronym{sxs}{SxS}{side-by-side}

\input{commands}

\title{Interactive Prompt Debugging with Sequence Salience}

\author{Ian Tenney,\textsuperscript{1}\Thanks{~~Equal contribution.} Ryan Mullins,\textsuperscript{1}\footnotemark[1] Bin Du,\textsuperscript{2} \\
  \textbf{Shree Pandya,\textsuperscript{2} Minsuk Kahng,\textsuperscript{1} and Lucas Dixon\textsuperscript{1}} \\
  \textsuperscript{1}Google Research \quad \textsuperscript{2}Google Cloud \\
  \texttt{\{iftenney, ryanmullins, dubin, shreepandya, kahng, ldixon\}@google.com} \\
}

\begin{document}
\maketitle
\begin{abstract}
We present \name{}, a visual tool for interactive prompt debugging with input salience methods. \name{} builds on widely used salience methods for text classification and single-token prediction, and extends this to a system tailored for debugging complex LLM prompts. Our system is well-suited for long texts, and expands on previous work by 1) providing controllable aggregation of token-level salience to the word, sentence, or paragraph level, making salience over long inputs tractable; and 2) supporting rapid iteration where practitioners can act on salience results, refine prompts, and run salience on the new output. We include case studies showing how \name{} can help practitioners work with several complex prompting strategies, including few-shot, chain-of-thought, and constitutional principles. \name{} is built on the Learning Interpretability Tool, an open-source platform for ML model visualizations, and code, notebooks, and tutorials are available at \projecturl{}.

\end{abstract}

\section{Introduction}
\label{sec:intro}

\begin{figure*}[!ht]
  \centering
  \includegraphics[width=\textwidth]{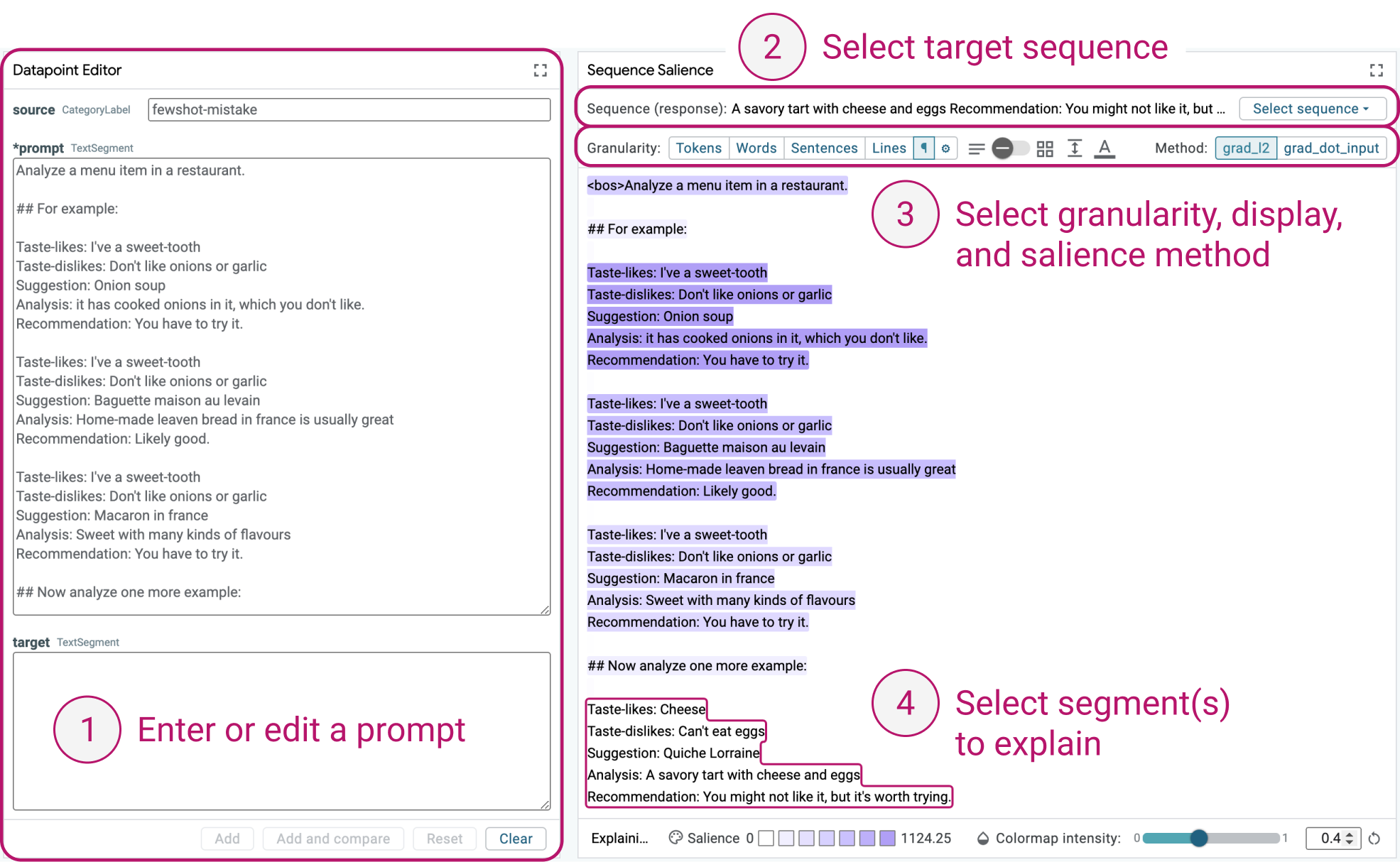}
  \caption{\name{} UI overview. The user can: (1) enter a prompt or edit an existing one, and optionally specify a target sequence to explain; (2) select a target sequence to explain, either a ground-truth sequence or an \gls{llm} generation; (3) control the selection granularity (tokens, words, sentences, lines, or paragraphs), visual display density, and a choice of salience methods (here, \texttt{grad\_l2} and \texttt{grad\_dot\_input}); and (4) select a segment, which triggers the system to compute salience with respect to that segment, showing the scores as a heatmap over preceding segments. Darker colors mean that segment is more influential or salient to the selected target. Shift-click can be used to select multiple segments, e.g., words comprising a phrase or clause.
  }
  \label{fig:hero}
\end{figure*}

\Glspl{llm} have become ubiquitous, presenting a text-to-text interface with the expressive power to solve a wide range of tasks that formerly required specialized solutions. Much of this usage is enabled by \emph{prompting} a pre-trained ``foundation'' model such as GPT-4 \cite{gpt4} or Gemini \cite{gemini}: providing carefully constructed inputs to elicit a behavior of interest, such as question answering or recommendation. Most practitioners interact with \glspl{llm} in a black-box manner \cite{ashtari2023discovery}: text goes in and the only feedback is the text that comes out. Even sophisticated prompt engineering tools \cite{jiang2022promptmaker,wu2022ai} operate in this setting, guiding developers only through external heuristics and design patterns.

We introduce a system, \name{} (Figure~\ref{fig:hero}), that uses \gls{is} methods to facilitate debugging and iteration on complex \gls{llm} prompts. Drawing on a long line of \gls{is} work for NLP tasks \cite{graddotinput,gradnorm,intgrad,lime,tenney2020language}, our system visualizes which parts of a prompt are important for a given output sequence.
While most \gls{is} methods do not \textit{causally} predict model behavior, they nonetheless provide useful heuristics which help guide a practitioner to what is likely to be important to the model \cite{ashtari2023discovery}. \name{} is fully interactive: a practitioner can make changes to the input and immediately see how the model responds, thereby enabling error detection and rapid prompt refinements to achieve the desired output.

Building on existing \gls{is} methods \cite{gradnorm,graddotinput}, \name{} extends their utility to modern \gls{llm} usage, which often deals with long texts and complex, semi-structured prompts.
In particular, while most salience methods operate natively at the token level, our system provides dynamic aggregation to coarser granularity---words, sentences, and paragraphs---significantly reducing the cognitive load on the practitioner who must interpret and act on these explanations. This enables, for instance, attribution to the logical set of few-shot examples or reasoning steps that more closely aligns with the developer's mental model. We explore several case studies in debugging prompt designs, including few-shot \cite{gpt2}, constitutional principles \cite{bai2022constitutional}, and chain-of-thought \cite{chainofthought}.

\name{} is built on the \Acrlong{lit} \citep[LIT;][]{tenney2020language}, a visual, interactive ML model debugging platform, which provides affordances for editing examples, comparing across models and examples, and managing sets of prompts.
\name{} is open-source and compatible with different \glspl{llm}, with code, notebooks, and tutorials available at \projecturl{}.

\section{System Design and Implementation}
\label{sec:system_design}
\name{} is implemented as a module on the \acrshort{lit} platform \cite{tenney2020language}. The frontend is written in TypeScript using the Lit (\url{http://lit.dev}) Web Components framework.
The Python backend hosts the \gls{llm} and supports three types of requests: \texttt{generate()}, which takes a prompt and produces one or more responses, \texttt{tokenize()} which returns a list of tokens for a given text, and \texttt{salience()} which takes a prompt, a target output sequence, and a target mask specifying which output token(s) to explain, and returns scores for the preceding tokens. The \texttt{salience()} function operates in ``teacher forcing'' mode, where the target sequence is given and gradients are taken with respect to the specified tokens. If the user changes the prompt, \texttt{generate()} is called and then \texttt{salience()} is called on the result.
As \gls{llm} calls can be expensive, a caching system and asynchronous requests are used to facilitate smooth, non-blocking interactions.

\name{} has been designed for left-to-right (``causal'') language models.
The implementation of gradient-based \gls{is} varies depending on the model class and runtime (e.g., PyTorch, TensorFlow, JAX). Models can be instrumented directly (see Appendix~\ref{sec:code_samples}),
or using an existing backend \cite[e.g.][]{inseq,miglani2023captum}.
To enable off-the-shelf use, we provide reference implementations for Gemma \cite{gemma}, Llama 2 \cite{llama2}, Mistral \cite{mistral}, and GPT-2 \cite{gpt2}.

\section{User Interface and Functionality}
\label{sec:ui_and_functionality}

\name{} is implemented as a browser-based user interface (Figure~\ref{fig:hero}). The user can enter a prompt or edit an existing one through the Datapoint Editor on the left side; alternatively they can use the Data Table (not shown) to select a pre-loaded example, such as from an evaluation set. The user then selects a sequence to explain, which can be either the response from the model or a pre-defined target. For a causal language model, each output token is a function of the preceding tokens---either from the prompt, or from earlier parts of the output. The user can click to select one or more output tokens (or aggregated segments), and salience will be computed and displayed as a heatmap over the text. Darker highlights correspond to higher salience scores, meaning those tokens are more important to the selected prediction target(s).

\paragraph{Prompt Editing}
Given their observations, the user may use the Datapoint Editor to edit their prompt, run the model, and re-compute salience; iteratively using this workflow can rapidly improve the prompt to achieve desired behavior (Section~\ref{sec:case_studies}). They can also provide an alternative \cite[contrastive; ][]{miller2019explanation} target sequence, in order to explore a ``\textit{what if}'' question about model behavior---this uses the same visualization, but shifts the explanation semantics from ``\textit{why did it predict this}'' to ``\textit{what would make it predict this},'' helping to identify parts of a prompt that should be reinforced or have the potential to cause problematic behavior.

\paragraph{Salience Methods}
The visualization is oriented around gradient-based salience methods, where a target span (one or more tokens) is specified and the method returns attributions simultaneously for all preceding tokens. Currently, we include \acrshort{gradnorm} \cite{gradnorm} and \acrshort{graddotinput} \cite{graddotinput}, as these methods are efficient to compute - an important consideration for interactive work with large models. Additional methods, such as \acrlong{ig} \cite{intgrad} or \acrshort{lime} \cite{lime}, can be implemented using a similar API. 

\paragraph{Granularity} 
Sense-making is easier when representations align with the users' mental model of the underlying data \cite{russell1993cost}.
Most salience methods return token-level attributions (for \glspl{llm} tokens are subwords from a tokenizer, e.g., SentencePiece; \citealt{sentencepiece}) and visualizations often use token-chips \cite{feng2019can,tenney2020language} to demarcate individual tokens and their scores. This can induce sense-making challenges compared to plain text as 1) tokens do not correspond directly to semantic units in natural language \cite{kaushal2022tokens}, affecting alignment with mental models, and 2) the visualization may insert formatting (e.g., line breaks, spacing, borders) which makes it difficult for the reader to scan and recognize different parts of the prompt.
This is exacerbated with long inputs, where there can be hundreds or thousands of tokens \cite{ding2024longrope}. 

To remedy this, \name{} provides two affordances. First, a flexible aggregation scheme where tokens can be grouped into more meaningful segments---words, sentences, lines, paragraphs, or custom units via a user-specified regular expression. Segments are mapped to the underlying token indices such that selecting a segment computes salience for the entire span. Salience scores are summed across segments and the color map is scaled accordingly.

Second, a running-text representation that remains faithful to the model's tokenization while rendering as close as possible to the original input text.
This may bias the user's visual attention \cite{wolfe2020visual}, as long segments will look more important than implied by their scores, as may aggregations that sum across many tokens. However, anecdotal evaluation across some use cases (Section~\ref{sec:case_studies}) suggests this is not a major issue: short segments (e.g. stopwords, punctuation) do not receive high salience scores, and the important segments are still easy to spot. Additionally, we provide a toggle to switch to traditional token chips for when a more precise but less-readable view is desired.

\paragraph{Further Affordances}
We provide a toggle to adjust the display density, which is helpful for working with longer text that may be challenging to fit on screen. 
The user can also adjust the colormap intensity, to emphasize the most salient segments or to increase contrast for diffuse attributions.
Finally, the \acrshort{lit} platform provides a \acrfull{sxs} functionality \cite{tenney2020language} which duplicates the visualization, allowing for flexible comparison across models or examples, as demonstrated in Section~\ref{sec:sxs_cot_example}.

\section{Case Studies}
\label{sec:case_studies}

Below, we give illustrative examples of using \name{} on common prompting scenarios.
These examples use Gemma,\footnote{\texttt{gemma\_instruct\_2b\_en} and \texttt{gemma\_instruct\_7b\_en} from \url{https://www.kaggle.com/models/google/gemma}} but the principles are general and apply to many common LLMs.

\subsection{Debugging Few-Shot Prompts}
\label{sec:few_shot_example}

Consider a developer building a new \gls{llm}-powered feature recommending dishes on a restaurant menu based on the \gls{llm}'s assessment of the user's tastes. Few-shot prompting \cite{gpt2} is an efficient way to implement this feature, as shown in the following prompt template.

\begin{lstlisting}[basicstyle=\ttfamily\scriptsize]
Analyze a menu item in a restaurant.

## For example:

Taste-likes: I've a sweet-tooth
Taste-dislikes: Don't like onions or garlic
Suggestion: Macaron in France
Analysis: Sweet with many kinds of flavours
Recommendation: You have to try it.

Taste-likes: I've a sweet-tooth
Taste-dislikes: Don't like onions or garlic
Suggestion: Onion soup
Analysis: It has cooked onions in it, which
you don't like.
Recommendation: You have to try it.

Taste-likes: I've a sweet-tooth
Taste-dislikes: Don't like onions or garlic
Suggestion: Baguette maison au levain
Analysis: Home-made leaven bread in France
is usually great
Recommendation: Likely good.

## Now analyse one more example:

Taste-likes: {{users-food-like}}
Taste-dislikes: {{users-food-dislike}}
Suggestion: {{menu-item}}
Analysis:
\end{lstlisting}

\begin{figure}[t!]
  \centering
  \includegraphics[width=\columnwidth]{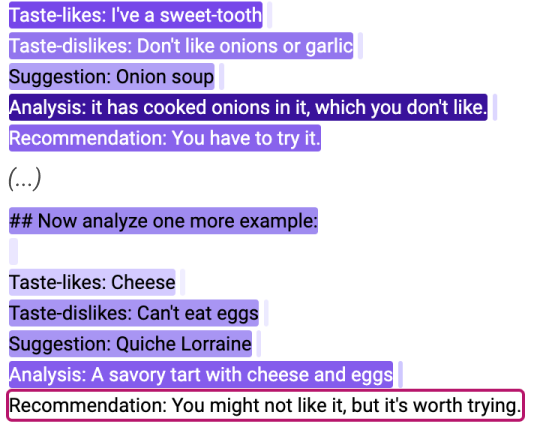}
  \caption{\name{} showing a sentence-level salience map for a user-selected sentence (``Recommendation...''). Here, the map suggests that an ``Analysis...'' sentence in the few-shot prompt is highly salient, but visual review shows it is followed by an incorrect recommendation.
  Some intervening text is hidden for space; for the full heatmap see Figure~\ref{fig:constitutions-full}.}
  \label{fig:few-shot}
\end{figure}

During testing, the developer observes a surprising generation on an example for a user that enjoys cheese, cannot eat eggs, and is considering the Quiche Lorraine item on a menu:

\begin{lstlisting}[basicstyle=\ttfamily\scriptsize]
Taste-likes: Cheese
Taste-dislikes: Can't eat eggs
Suggestion: Quiche Lorraine
Analysis: A savoury tart with cheese and eggs
Recommendation: You might not like it, but it's
worth trying.
\end{lstlisting}

The developer wonders why the model recommends an item that contains an ingredient (eggs) the user cannot eat, and uses sentence-level Sequence Salience to better understand how the few-shot examples in the prompt contributed to this output. They select the ``Recommendation'' sentence from the model's response, and the Sequence Salience module computes and visualizes salience as a heatmap, shown in Figure~\ref{fig:few-shot}.

As the developer scans up through the few-shot examples, they notice that the ``Analysis'' line of the second example is quite strong (deep purple), suggesting the model relied heavily on that analysis in this example, but the ``Recommendation'' line that follows is incorrect: it says ``You have to try it!'' when it should say something along the lines of ``Not recommended''. The developer realizes this is likely a copy-paste error when creating the prompt template, fixes the few-shot example, and re-runs the test to verify the fix.

\subsection{Assessing Constitutional Principles}
\label{sec:constitution_example}

Continuing with the above scenario, suppose the developer wants to experiment with adding principles \cite[also known as \textit{constitutions};][]{bai2022constitutional,petridis2024constitutionmaker} to the prompt to help guide the model's generation. They directly edit the prompt in the Datapoint Editor, adding two principles before the few-shot examples:

\begin{lstlisting}[basicstyle=\ttfamily\scriptsize]
* The analysis should be brief and to the point.
* The analysis and recommendation should both be 
clear about the suitability for someone with a 
specified dietary restriction.
\end{lstlisting}

\begin{figure}[!ht]
  \centering
  \includegraphics[width=\columnwidth]{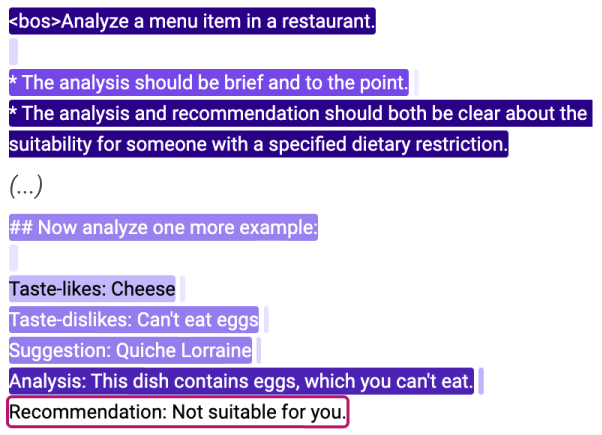}
  \caption{\name{} highlighting the influence of the constitutional principles the developer added to the beginning of the prompt, relative to the selected sentence (``Recommendation...''), helping to assess the effectiveness of prompt iterations. Some text is hidden for space; for the full heatmap see Figure~\ref{fig:constitutions-full}.}
  \label{fig:constitutions}
\end{figure}

When the example is edited, \name{} runs generation automatically, allowing the user to inspect the new response (see Figure~\ref{fig:constitutions}) to get a sense of what changed. Looking again at sentence-level salience, the recommendation ``Not suitable for you'' is strongly and equally influenced by the second principle (``...be clear about the suitability for someone with a specified dietary restriction'') and the explanatory analysis statement generated by the model, as desired.

\subsection{Comparing Behavior Side-by-Side}
\label{sec:sxs_cot_example}

\begin{figure*}[!ht]
  \centering
  \includegraphics[width=0.93\textwidth]{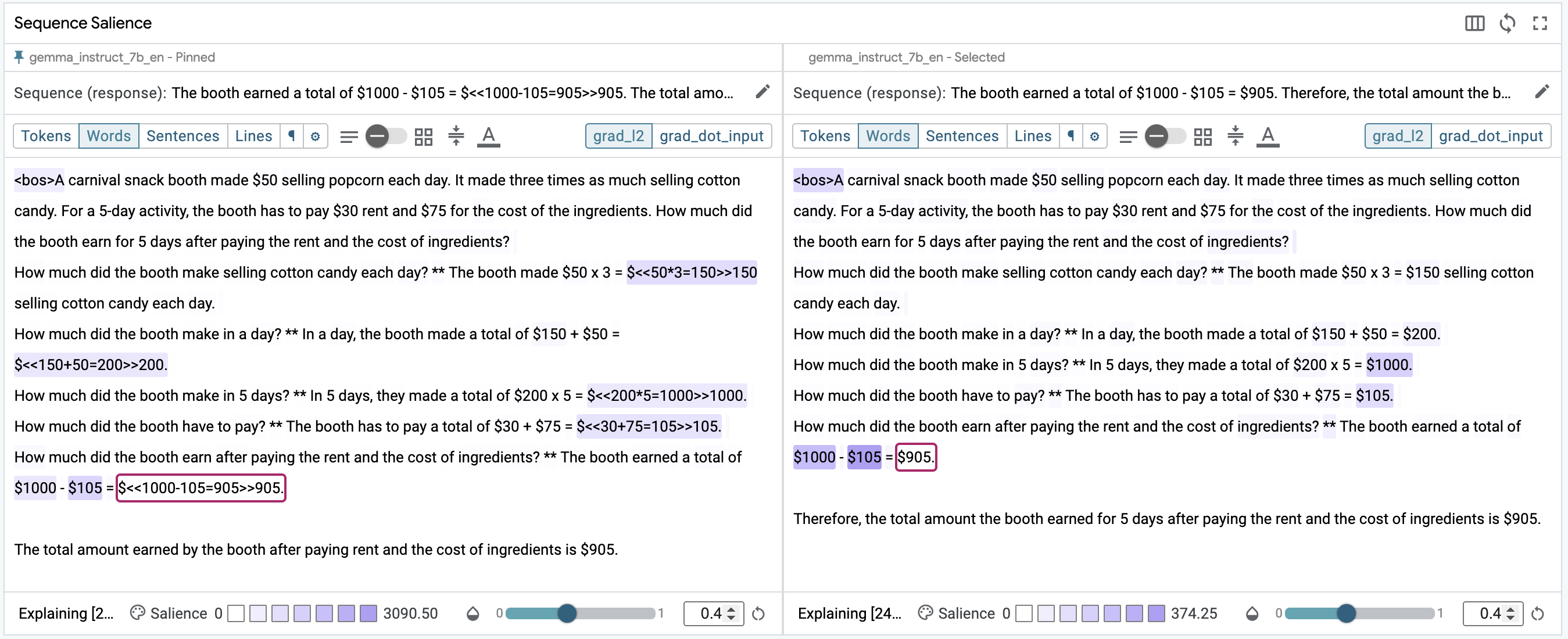} 
  \caption{Side-by-side, sentence-level \name{} maps comparing results for two variants of a GSM8K example. 
  The left side shows the original example and shows a diffuse salience map across the numerical values. The right side modifies the prompt to remove the calculation annotations, yielding a more focused salience map over the operands and relevant answers, which in turn reveals issues with specific arithmetic calculations.}
  \label{fig:uc3_sxs}
\end{figure*}

\gls{lit} provides a \gls{sxs} comparison feature, accessed by ``pinning'' a datapoint in the main toolbar or the Data Table. Here, we consider how \gls{sxs} comparison with \name{} can be used to debug model output against ground truth and counterfactual examples.

GSM8K \cite{gsm8k} is a dataset of math problems used to evaluate recent 
\glspl{llm} \cite{gemma,llama2,mistral}, typically using a
\gls{cot} prompt from the conventional form of \citet{chainofthought}. A developer might expect a model to perform similarly or better on the dataset's Socratic form, which encodes the \gls{cot} as a series of question-answer pairs in the prompt. However, for the example in Figure~\ref{fig:uc3_sxs}, this is not the case.

In \gls{sxs} mode, the \name{} visualization is duplicated, showing the ``pinned'' example on the left and the selected example on the right. The user controls which target sequence is shown for each example, useful for comparing ground truth against a generated response.
A quick look shows that the model is setting up the problem with the correct operands, but gets the computation wrong. Word-level salience results for both the ground truth (not shown) and generated response (Figure \ref{fig:uc3_sxs} left) show uniform attribution across all prior answers. The developer wonders if this is impacting performance, and hypothesizes that removing GSM8K's calculation annotations \citep{gsm8k} might enable correct calculations.

After revising the selected example, the developer compares original and revised examples in \gls{sxs} mode (Figure~\ref{fig:uc3_sxs}). Word-level salience shows that the model attends much more strongly to the operands in the revised example (Figure~\ref{fig:uc3_sxs} right), suggesting that removing the calculation annotations does focus the model. However, the model still fails to compute the correct value. At this point, the developer believes that this model, like many other \glspl{llm}, struggles with arithmetic calculations, and that tool use approaches \cite{toolformer} may be required to improve performance.

\section{Related Work}
\label{sec:related_work}

\paragraph{Explanations for LLMs}
The necessity and desiderata of explanations for ML behavior have been extensively studied \cite{doshi2017,intgrad,Kaur2020,miller2019explanation,mueller2019explanation}. However, these foundations were established in the context of smaller models on simpler tasks, such as classification.
\glspl{llm} present several novel challenges to prior interpretability methods, primarily due to their scale---the number of parameters \cite{gpt4}, the context window size \cite{ding2024longrope}, and the diversity of supported tasks \cite{gemini}---and due to the subjective evaluation of their outputs. Concretely, assessing explanation sufficiency for \glspl{llm} requires more cognitive effort to judge both the acceptability of the model's output 
and the explanation relative to the input and output features. 
This is amplified by the natural domain of salience methods (sub-word tokens) compared to the general semantic units (words, phrases, sentences) that humans are used to reasoning about \cite{kaushal2022tokens}.

\paragraph{Gradient-based Text Salience}
\Acrfull{is} methods are a class of explanations used to estimate the sensitivity of a model's prediction relative to the presence of specific input features \cite{camburu2020struggles,intgrad}; for text-to-text LLMs, these are typically the tokenized text \cite{sentencepiece}.
This work focuses on the generalization of \emph{gradient-based methods} \cite{gradnorm,graddotinput,intgrad}, though many aspects are also applicable to ablation-based methods \cite{lime,shap}.
As \glspl{llm} are expensive to evaluate, we focus on \acrshort{gradnorm} and \acrshort{graddotinput} for their computational efficiency, as only a single inference call is required to compute salience for an example.
Prior work has assessed the utility of these methods for model analysis with mixed results. \citet{bastings2021will} found that efficacy depends on the model architecture and task, and recommended \acrshort{gradnorm} as a good default. \citet{madsen2023faithfulness} also measures faithfulness and finds \acrshort{gradnorm} to perform comparably to more expensive methods on a range of classification tasks.
Despite this potential for misalignment, \citet{ashtari2023discovery} found that humans could make sense of results from these methods in a task-driven scenario.

\paragraph{Visualizing Salience for Interpretation}
\Acrlong{is} results are often visualized as a heatmap over a representation of the input and output features \cite{feng2019can}. Prior works such as LIME \cite{lime}, Captum \cite{kokhlikyan2020captum,miglani2023captum}, and \gls{lit} \cite{tenney2020language} use a heatmap over tokens to represent feature importance for text classification, where salience scores are mapped to the text's background color; closest to our work is ECCO \cite{ecco} which also presents a running-text view, but focuses on fine-grained token-level information.
Applications to seq2seq tasks have also used matrix-style visualizations where scores are arranged in a grid of input and output tokens \cite{intgrad,inseq}.
\citet{seq2seqvis} also extends this approach using compound visualizations that combine elements from heatmaps, bar charts, and Sankey diagrams.
Additionally, LMDiff \cite{strobelt2021lmdiff} and \gls{lit} use linked, coordinated visualization techniques \cite{dork2008visgets} to enable side-by-side comparison, among other affordances.
Finally, LLMCheckup \cite{llmcheckup} proposes a dialogue-based UI that uses an LLM to parse and format explanations.

\section{Conclusions and Future Work}
\label{sec:discussion_future_work}

We introduce \name{}, an interactive, visual tool for prompt debugging, using salience methods to understand what parts of the input are most relevant to the model. \name{} is designed to work well with longer text and with complex prompting strategies such as few-shot, chain-of-thought, and constitutional principles. By aggregating to the word, sentence, line, or paragraph level, \name{} reduces the cognitive overhead of interpreting the salience scores and makes explanations more directly actionable. By making it easy to edit the example and validate the effect on the model, we enable developers to rapidly iterate on and improve LLM prompts.

The current iteration of \name{} supports \acrshort{gradnorm} and \acrshort{graddotinput} salience methods, works out-of-the-box with Gemma, Llama 2, Mistral, and GPT-2 models (TensorFlow or PyTorch), and is easily extensible to new models and salience methods.
By making this available, we hope to increase adoption of these methods for prompt debugging, as well as to foster development of improved salience methods for LLMs. For example, by better understanding the cases in which salience accurately predicts changes in model behavior \cite{faithfulxnlpsurvey}, we can develop more causally-robust methods. Moreover, the visualization can be extended to support additional modalities, such as salience-guided counterfactuals \cite{hotflip,wallace2019universal} or contrastive explanations \cite{jacovi2021contrastive,yin2022interpreting}.

\section*{Acknowledgements}
We thank Oscar Wahltinez, Matthew Watson, Cibi Arjun, Ludovic Peran, Mark Collins, as well as the members of Google's People + AI Research (PAIR) team for many helpful discussions, support, and feedback.

\bibliography{main}

\clearpage
\appendix
\setcounter{figure}{0} \renewcommand{\thefigure}{A.\arabic{figure}}
\onecolumn

\section{Full-text Examples}
\label{sec:full_examples}

These show the full text and salience heatmap for the examples from Figure~\ref{fig:few-shot} and Figure~\ref{fig:constitutions}, including intervening few-shot examples.

\begin{figure*}[!ht]
  \centering
  \begin{subfigure}[t]{0.48\textwidth}
    \includegraphics[width=\textwidth]{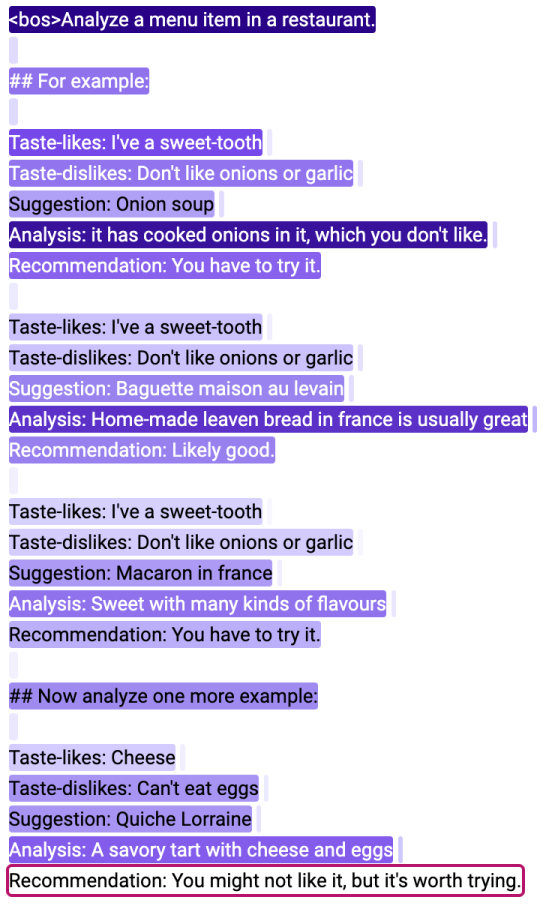}
    \caption{\name{} showing a sentence-level salience map for a user-selected sentence (``Recommendation...''). Here, the map suggests that an ``Analysis...'' sentence in the few-shot prompt is highly salient, but visual review shows it is followed by an incorrect recommendation, helping users find errors in prompts.}
    \label{fig:few-shot-full}
  \end{subfigure}\hfill%
  ~
  \begin{subfigure}[t]{0.48\textwidth}
    \includegraphics[width=\textwidth]{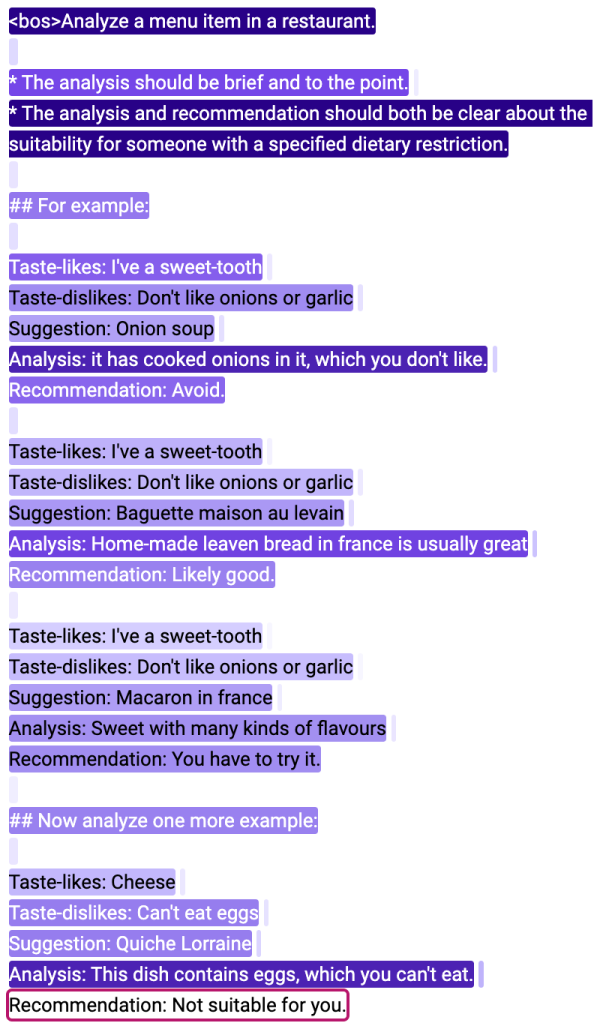}
    \caption{\name{} highlighting the influence of the constitutional principles the developer added to the beginning of the prompt, relative to the selected sentence (``Recommendation...''), helping to assess the effectiveness of prompt iterations.}
    \label{fig:constitutions-full}
  \end{subfigure}
  \caption{Full text heatmaps for examples from Figure~\ref{fig:few-shot} and Figure~\ref{fig:constitutions}.}
\end{figure*}

\clearpage
\section{Code Examples}
\label{sec:code_samples}

\setcounter{figure}{0} \renewcommand{\thefigure}{B.\arabic{figure}}

To use gradient-based salience methods, it is necessary to instrument the model and provide access to some internal tensors. A simple way to implement this is to add a callback API similar to Garcon \cite{garcon}. An idealized example for a decoder-only Transformer model is shown in Figure~\ref{fig:model-code-example}:

\begin{figure*}[h!]
    \centering
    \usemintedstyle{vs}
\begin{minted}{python}
 def score(self, input_ids, padding_mask, target_ids,
           layer_intercept_fn = lambda x, i: x):
   """Compute per-token loss, with optional layer hooks."""
   x = self.embedding_layer(input_ids)
   # Embeddings as layer 0
   x = layer_intercept_fn(x, 0)
   for i, layer in enumerate(self.transformer_layers):
     x = layer(x, padding_mask=padding_mask)
     # Transformer layers from 1 to N
     x = layer_intercept_fn(x, i+1)

   logits = self.output_layer(x)

   loss_fn = sparse_categorical_crossentropy(
       from_logits=True, reduction="none"
   )
   # <float>[batch_size, num_tokens]
   per_token_loss = loss_fn(target_ids, logits)
   return per_token_loss
\end{minted}
    \vspace{-1em}
    \caption{Example model code, instrumented to accept a \texttt{layer\_intercept\_fn} which can be used to implement a variety of interpretability methods, including gradient-based salience. A full implementation for Gemma is available at \url{https://github.com/keras-team/keras-nlp/blob/v0.8.2/keras_nlp/models/gemma/gemma_causal_lm.py\#L313}.}
    \label{fig:model-code-example}
\end{figure*}

Using this interface, it is straightforward to access intermediate tensors and compute salience. This example below in Figure \ref{fig:salience-code-example} implements \acrshort{gradnorm}:

\begin{figure*}[h!]
    \centering
    \usemintedstyle{vs}
\begin{minted}{python}
embs = None
with tf.GradientTape() as tape:

  def watch_embs(x, layer_index):
    """Watch embeddings and store a reference to them for a later .gradient() call."""
    global embs
    if layer_index == 0:
      tape.watch(x)  # watch for gradients
      embs = x       # store a reference
    return x

  per_token_loss = model.score(input_ids, padding_mask, target_ids,
                               layer_intercept_fn=watch_embs)
  # Apply target mask to select which output positions we're interested in gradients for.
  masked_loss = per_token_loss * target_mask

# <tf.float32>[batch_size, num_tokens, emb_dim]
grads = tape.gradient(masked_loss, embs)
# <tf.float32>[batch_size, num_tokens]
grad_l2 = tf.norm(grads, axis=2)
\end{minted}
    \vspace{-1em}
    \caption{Example salience code to implement \acrshort{gradnorm} in TensorFlow using the callback API shown in Figure~\ref{fig:model-code-example}. A full implementation for KerasNLP models running on TensorFlow or PyTorch can be found at \url{https://github.com/PAIR-code/lit/blob/v1.1.1/lit_nlp/examples/models/instrumented_keras_lms.py\#L256}.}
    \label{fig:salience-code-example}
\end{figure*}

The full implementation requires some additional code to handle batching and preprocessing, as well as to implement the \texttt{generate()} and \texttt{tokenize()} functions. Additional information on the APIs can be found in the \gls{lit} documentation at \url{https://pair-code.github.io/lit/documentation/}, and the full code for \name{} can be found at \projecturl{}.

\end{document}

%% file: commands.tex
\newcommand{\name}[1]{Sequence Salience}

\newcommand{\projecturl}[1]{\url{http://goo.gle/sequence-salience}}